\documentclass[runningheads]{llncs}
\usepackage[T1]{fontenc}
\usepackage{graphicx}
\usepackage{amsmath}
\usepackage{amssymb}
\usepackage{multirow}
\usepackage{rotating}
\usepackage{adjustbox}

\begin{document}

\title{Time-Conditioned and Multi-Time Survival Prediction from 2D PET/CT Projections in Lung Cancer}
\titlerunning{Time-Conditioned and Multi-Time Survival Prediction}

%

\author{Ashish Chauhan\inst{1}\orcidID{0009-0009-3050-1264} \and
Sambit Tarai\inst{1,2}\orcidID{0000-0002-5550-3575} \and
Elin Lundström\inst{1}\orcidID{0000-0003-2955-4958} \and
Johan Öfverstedt\inst{1}\orcidID{0000-0003-0253-9037} \and
Håkan Ahlström\inst{1,3}\orcidID{0000-0002-8701-969X} \and
Joel Kullberg\inst{1,3,4}\orcidID{0000-0001-8205-7569}
}


%
\authorrunning{A. Chauhan et al.}

%
\institute{Radiology, Department of Surgical Sciences, Uppsala University, Uppsala, Sweden \\
\email{\{ashish.chauhan, sambit.tarai, elin.lundstrom, johan.ofverstedt, hakan.ahlstrom, joel.kullberg\}@uu.se} \and
National Academic Infrastructure for Supercomputing (NAISS), Linköping University, Linköping, Sweden
\and
Antaros Medical, Mölndal, Sweden \and
SciLifeLab, Uppsala University, Uppsala, Sweden\\
}
\maketitle              

\begin{abstract}

Accurate prediction of overall survival (OS) from positron emission tomography/computed tomography (PET/CT) can support personalized treatment and follow-up strategies in oncology. However, the impact of temporal modeling on imaging-based survival prediction remains insufficiently explored. We investigate how different temporal formulations influence survival prediction by developing two complementary approaches: Attention-guided Time-Conditioned Survival (ATCS) and Multi-Time Survival (MTS).
We retrospectively analyzed pre-treatment PET/CT images from 848 patients with non-small cell lung cancer (NSCLC), including 556 for model development and 292 for held-out testing. 
A previously proposed Time-Conditioned Survival (TCS) model was used as a baseline. Models were trained using 5-fold cross-validation and evaluated on the test set using time-dependent area under the curve (AUC) at 6-month intervals from 0.5 to 5 years.
Both ATCS and MTS outperformed the baseline TCS model, achieving mean AUCs of 0.794 and 0.793, respectively, compared to 0.767. ATCS performed better at earlier time points (0.5–3 years), whereas MTS performed better at later intervals (3.5–5 years). Combining tumor-specific and tissue-wise PET/CT features improved performance over either input alone. Finer temporal discretization improved short-term prediction, while coarser intervals provided more stable long-term estimates.
These findings demonstrate that temporal modeling and input design influence PET/CT-based survival prediction. The proposed approaches enable time-specific survival estimation from pre-treatment imaging and may support improved risk stratification and clinical decision-making.

\end{abstract}

\section{Introduction}

Lung cancer ranks among the leading causes of cancer-related mortality worldwide, with non-small cell lung cancer (NSCLC) comprising over 80\% of all diagnosed cases \cite{bray2024global,hendriks2024non}. Due to the absence of early symptoms and inadequate screening, diagnosis often occurs at advanced stages, leading to poor patient outcomes \cite{gridelli2015non}. Prognosis in NSCLC is commonly inferred from a combination of clinical assessments, pathological findings, and radiological imaging results \cite{garinet2022updated}. The tumor, node, and metastasis (TNM) staging system, based on tumor characteristics such as size and spread, lymph node involvement, and distant metastasis, plays a key role in guiding treatment decisions and estimating prognosis \cite{lababede2018eighth,woodard2016lung}. Treatment strategies may include surgery, radiotherapy, chemotherapy, immunotherapy, or targeted treatments for tumors with specific genetic alterations.

Medical imaging with fluorodeoxyglucose positron emission tomography/computed tomography (FDG-PET/CT) plays a crucial role in the management of NSCLC, offering both anatomical and metabolic insights. Beyond staging, the ability of FDG-PET/CT to quantify tumor metabolism serves as an indicator of aggressiveness and treatment response  \cite{khiewvan2016role}. 
Medical imaging plays a central role in oncology, and deep learning methods have shown promise in improving prognostic modeling and patient risk stratification from imaging data.

In prior work \cite{tarai2026time}, a time-conditioned deep learning framework was explored, here referred to as Time-Conditioned Survival (TCS), that combines tissue-wise PET/CT projections with follow-up time to estimate overall survival (OS) as a continuous function of time. This approach demonstrated improved performance over fixed-interval prediction strategies and enabled patient risk stratification. However, the continuous-time formulation required repeated evaluation across time and was therefore computationally more demanding compared to fixed-interval approaches. In addition, it offered limited flexibility in integrating temporal information with imaging features, leaving scope for further architectural and training refinements. 

The present study investigates alternative temporal modeling strategies for PET/CT-based survival prediction in NSCLC. Specifically, we explore an Attention-guided Time-Conditioned Survival (ATCS) approach alongside a complementary Multi-Time Survival (MTS) framework for discrete OS prediction. We analyze how different temporal representations influence survival modeling and examine the role of input design and interpretability in improving prognostic performance. More broadly, this work aims to advance flexible and clinically interpretable models that account for the temporal complexity of cancer outcomes. An overview of the proposed frameworks is provided in Fig. ~\ref{fig:overview_models}.

The main contributions of this work are as follows:
\begin{itemize}
\item Architectural and training refinements of time-conditioned survival modeling framework.
\item Development and evaluation of a multi-time survival modeling framework for discrete OS prediction across different temporal resolutions.
\item Systematic assessment of the impact of input configurations, including tissue-wise PET/CT projections, tumor-specific channels, and their combinations, on multi-time survival modeling performance.
\item Comparative analysis of time-conditioned and multi-time survival modeling strategies for OS prediction in NSCLC, complemented by cohort-level saliency analysis for interpretability.
\end{itemize}

\section{Related works}

\subsection{Traditional and Neural Network-Based Survival Models}

Survival analysis has historically relied on the Cox proportional hazards model (Cox, 1972), a semi-parametric regression approach that links covariates to hazard rates under the assumption of proportional hazards \cite{hespanhol1995survival}. 
While widely used, its linear formulation limits its ability to capture complex relationships.

To address this limitation, Katzman \textit{et al.}~\cite{katzman2018deepsurv} proposed DeepSurv, which replaced the Cox linear predictor with a deep neural network, enabling nonlinear modeling. Although effective, DeepSurv retained the proportional hazards assumption. Kvamme \textit{et al.}~\cite{kvamme2019time} further relaxed this constraint by incorporating time as a covariate, allowing risk factors to vary dynamically. Their approach, combined with a loss inspired by nested case-control studies, improved flexibility across proportional and non-proportional settings. In parallel, discrete-time survival models reformulate the problem by partitioning follow-up into predefined intervals and estimating event probabilities within each interval. In this formulation, survival prediction is often cast as a sequence of binary classification tasks over time intervals, enabling regression via classification and allowing flexible modeling of time-dependent risk. Methods such as DeepHit \cite{lee2018deephit} model the event-time distribution on a discretized time grid, enabling flexible modeling beyond proportional hazards assumptions. These continuous and discrete temporal parameterizations represent distinct strategies for neural survival modeling.

\subsection{Deep Learning in Medical Imaging Prognostics and PET/CT Applications}

Convolutional neural networks (CNNs) have demonstrated strong potential for extracting prognostic features from medical imaging. Prior studies have shown that CNN-based models can predict clinical outcomes such as OS from CT images \cite{chen2021deep,diamant2019deep}, while deep learning-based radiomics approaches have outperformed conventional radiomics in survival prediction tasks \cite{huynh2023head}. Multimodal imaging strategies have further improved performance by integrating anatomical and functional information. Amini \textit{et al.}~\cite{amini2021multi} demonstrated that combining FDG-PET and CT features enhances survival prediction in NSCLC.

FDG-PET/CT has been widely explored for outcome prediction, particularly for deriving prognostic biomarkers. Tarai \textit{et al.}~\cite{tarai2024prediction} proposed tissue-wise multichannel PET/CT projections for estimating total metabolic tumor volume (TMTV), a known prognostic factor for OS \cite{girum202218f}. Similar projection-based approaches have also been applied in other cancer types \cite{ferrandez2023artificial}.

\subsection{Interpretability}

Interpretability is an important complement to predictive modeling in medical imaging, particularly for clinical decision support. Various saliency and attribution methods have been proposed for CNNs, among which Gradient-weighted Class Activation Mapping (Grad-CAM) \cite{selvaraju2017grad} is widely used to visualize spatial regions contributing to model predictions. Grad-CAM generates class-discriminative saliency maps using gradient information and feature activations and can be applied without architectural modification.

Beyond individual-level visualization, cohort-level 
analysis has been explored in whole-body PET/CT imaging. Jönsson \textit{et al.}~\cite{jonsson2023spatial,jonsson2022image} introduced a registration framework for aligning images into a common template space, enabling voxel-wise population-level analysis. Tarai \textit{et al.}~\cite{tarai2024prediction} extended this approach to cohort-level saliency analysis, motivating the strategy adopted in this work.

\begin{figure}[htbp]
  \centering 
  \includegraphics[width=0.85\textwidth]
  {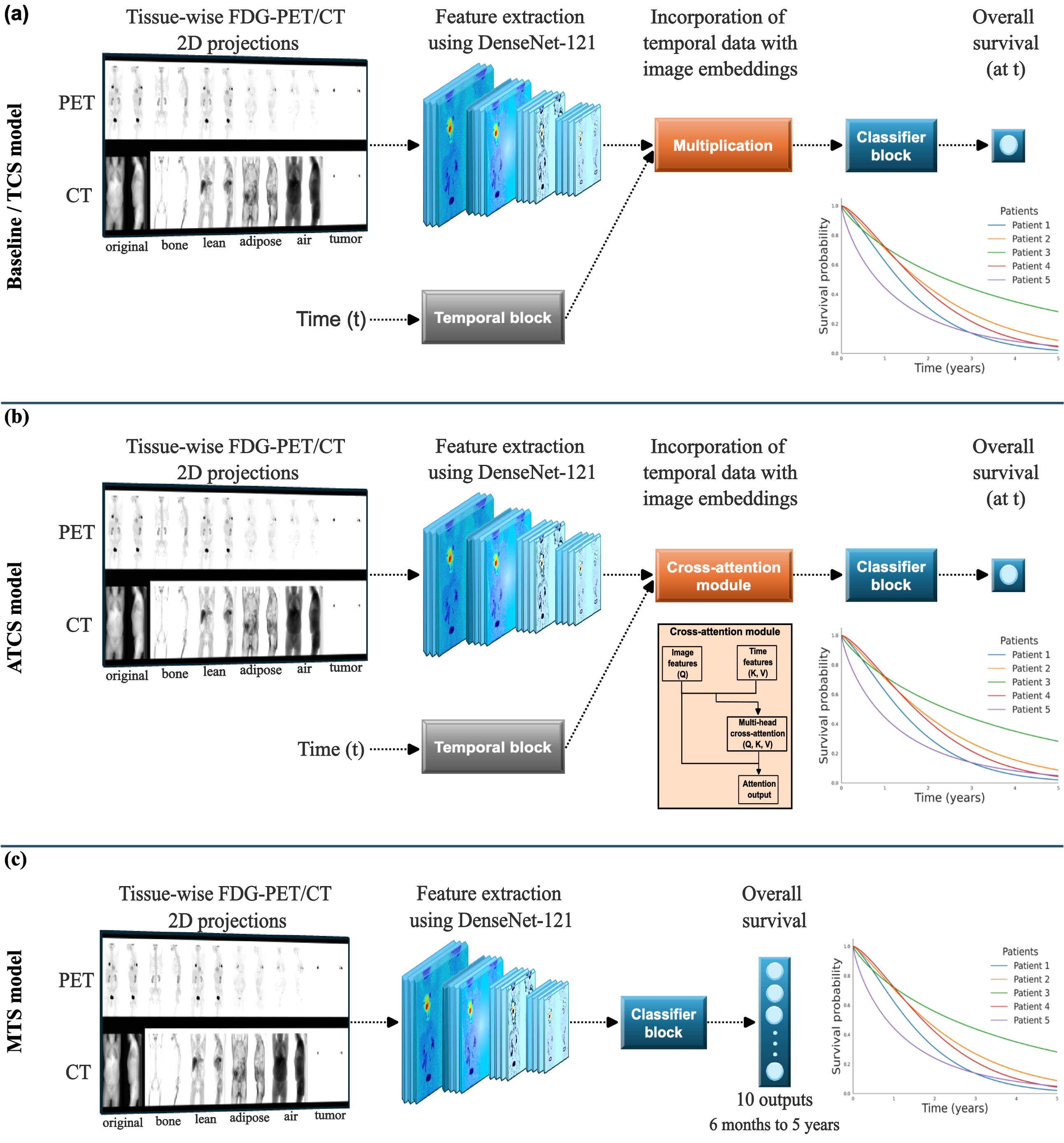} 
  \caption{Overview of the proposed frameworks for overall survival (OS) prediction from tissue-wise 2D PET/CT projections using a DenseNet-121 backbone. (a) Baseline/time-conditioned survival (TCS), where OS is modeled as a function of follow-up time. (b) Attention-guided time-conditioned survival (ATCS), which extends TCS via attention-based fusion of imaging features and temporal information. (c) Multi-time survival (MTS), which predicts survival probabilities at fixed 6-month intervals (from 6 months to 5 years) for discrete-time OS estimation and survival curve generation.}
  \label{fig:overview_models}
\end{figure}

\section{Methods}

\subsection{Dataset}

The dataset used in this study is from Uppsala-Umeå comprehensive consortium (U-CAN) \cite{glimelius2018u}. This subset consists of whole-body FDG-PET/CT images of NSCLC patients. For this study, we selected baseline images acquired prior to treatment. Patients with fewer than 90 days of follow-up were excluded, and survival times were truncated at 5 years. The final cohort comprised 848 patients, of whom 556 were used for model development and internal evaluation through 5-fold cross-validation, and a held-out test set of 292 patients was used for final performance assessment.

All PET/CT images were resampled to uniform voxel spacing of (2.04 × 2.04 × 3.00) mm³ with linear interpolation and PET images were converted to standardized uptake values (SUV) \cite{kinahan2010positron}. Tumor segmentation proposals for the training set were generated using a U-Net model pretrained on the AutoPET dataset \cite{tarai2024improved}, then refined by a nuclear medicine and radiology specialist (10+ years of experience). Expert-corrected annotations were subsequently used to fine-tune the model before generating automatic segmentations for the test set. Ethical approval was obtained from the Swedish Ethical Review Authority (Dnr 2023-02312-02).

\subsection{Tissue-wise FDG-PET/CT projections and tumor slices}

The PET and CT volumes were separated into tissue-specific channels based on CT Hounsfield units, following previously established thresholds as described in \cite{tarai2024prediction}. Voxels were classified as bone, lean soft tissue, adipose tissue, and air. 
For each tissue type, maximum intensity projections (MIPs) and average intensity projections (AIPs) were generated for PET and CT, respectively, along the coronal and sagittal planes, providing a computationally efficient 2D representation of the 3D data while preserving key anatomical and physiological information (see Fig. \ref{fig:overview_models}). 
Tumor segmentation masks were applied to generate tumor-specific projections from both PET and CT volumes. These included MIPs from tumor-masked PET images, AIPs from tumor-masked CT images, and sum intensity projections (SIPs) derived from the tumor segmentation.
In addition, representative tumor slices were extracted from the PET and CT volumes for each patient. For each tumor region, the coronal and sagittal slices with the largest tumor cross-sectional area were selected. When multiple tumor regions were present, slices were overlaid in ascending order of tumor area, such that larger tumor regions were preserved in cases of spatial overlap. This process yielded a single tumor slice collage per patient.
In total, the input consisted of tissue-wise PET/CT projections derived from original and tissue-specific regions (bone, lean, adipose, and air), along with tumor-masked projections and tumor segmentation–based projections. These components are referred to as tissue-wise projections ($Tissues$), tumor-masked projections ($T_{masked}$), tumor segmentation projections ($T_{seg}$), and tumor slices ($T_{slice}$), as used in subsequent experiments.

\subsection{Models}

The baseline model (TCS) serves as a reference for evaluating subsequent enhancements, including the ATCS and MTS models. All models were designed for OS prediction over a 5-year period, evaluated at 6-month intervals.

\subsubsection{Time-Conditioned Survival (TCS)}  

The TCS model (see Fig.~\ref{fig:overview_models}), previously proposed in \cite{tarai2026time}, employs a ResNet-50 backbone to integrate temporal information with image features extracted from tissue-wise multichannel PET/CT projections. 
A dynamic time-sampling strategy was used during training, where random time points between 1 month and 5 years (sampled at 1-month intervals) were generated for each patient. 
This enables the model to learn relationships between imaging features and temporal progression for OS prediction.

For each patient, the time interval was defined as the period from the baseline images to the clinical follow-up date for survivors, and as two segments, from baseline to death and from death to follow-up for deceased patients.
To address the limited number of early events (deaths), earlier time points were oversampled during training. 
Censoring was handled by including only sampled follow-up times up to the censoring time for alive patients, while for deceased patients the event time was used to define survival status at each sampled time point.
A single network was trained to predict OS at pre-specified time points from 6 months to 5 years.

For this baseline experiment, the DenseNet-121 backbone was selected based on empirical evaluation, as it consistently outperformed alternative architectures for the TCS framework (see Supplementary Table \ref{tab:suppl_P1_arch_ablation}).

\subsubsection{Attention-Guided Time-Conditioned Survival (ATCS)}

Building on the TCS model, the ATCS model (see Fig. \ref{fig:overview_models}) was developed to refine the integration of temporal information with imaging features and to improve training stability. The primary architectural modification in the ATCS model was the replacement of the element-wise feature-time interaction used in the TCS model with a cross attention module \cite{gheini2021cross}. In this formulation, imaging features extracted by the backbone were fused with a learned time embedding through an attention mechanism to produce time-conditioned feature representations.

In addition to architectural changes, the ATCS model incorporated several training refinements, including gradient accumulation to achieve larger effective batch sizes under memory constraints, a learning-rate warmup scheduler, and a parameterized loss-weighting strategy in which task weights were learned during training via trainable log-variance parameters (see Supplementary Section \ref{section:suppl_ATCS_training_strat}).

A vision transformer (ViT)-based variant of the ATCS model \cite{dosovitskiy2020image} was also evaluated; however, DenseNet-121 again achieved superior performance (see Supplementary Section \ref{section:suppl_additional_res}).

\subsubsection{Multi-Time Survival (MTS)}

The MTS model (see Fig. \ref{fig:overview_models}) was developed to predict survival probabilities at multiple time points. Similar to the previous approaches, it employed a DenseNet-121 backbone combined with gradient accumulation and a warmup scheduler (see Supplementary Section \ref{section:suppl_MTS_training_strat}).
Whereas the earlier ATCS model incorporated temporal information by embedding time into the image features, it still lacked a constraint to ensure realistic survival dynamics, specifically, penalizing non-monotonic behavior in survival probabilities. Since survival probability is inherently monotonic (non-increasing) over time, we enforced this property by defining each output node as the product of its predecessors. 
For example, the probability from output node 1 ($O_{1}$), corresponding to the first time interval, was multiplied with that of output node 2 ($O_{2}$) to obtain the final probability of output node 2 ($O_{2}^{\text{final}}$). Extending this, the final probability for the $m$th output node, corresponding to the last time interval, was computed as

\begin{equation} \label{eq:MTS_out}
O_{m}^{\text{final}} = O_{1} \times O_{2} \times \cdots \times O_{m},
\end{equation}

resulting in a monotonically decreasing survival probability curve.
Censoring was incorporated by computing the loss only for discrete time points up to the censoring time for alive patients, whereas for deceased patients, survival status was defined according to the observed event time across the evaluated intervals.

\subsection{Loss function}

For the TCS model, the training objective combines a focal loss ($\mathcal{L}_{\text{focal}}$), derived from binary cross-entropy, with a survival consistency loss ($\mathcal{L}_{\text{SCL}}$) that penalizes non-monotonic survival probabilities over time. 
Focal loss was selected to address the class imbalance problem in survival labels, which is particularly pronounced at early and late follow-up time points. 
The total loss is defined as

\begin{equation}
\mathcal{L}_{\text{total}} = \mathcal{L}_{\text{focal}} + \lambda \cdot \mathcal{L}_{\text{SCL}},
\label{eq:P1_total_loss}
\end{equation}

where $\lambda$ controls the relative contribution of the consistency term and was set to $\lambda = 1$ in this study.

For the ATCS model, the fixed weighting scheme was replaced with an adaptive loss-weighting strategy based on task-dependent uncertainty, following the approach of Kendall \textit{et al.}~\cite{kendall2018multi}. In this formulation, the relative contributions of the focal loss and the survival consistency loss are learned during training through two trainable log-variance parameters, $\log \sigma^{2}_{\text{cls}}$ and $\log \sigma^{2}_{\text{aux}}$, associated with the classification and auxiliary consistency tasks, respectively. The resulting objective function is given by

\begin{equation}
\mathcal{L}_{\text{total}} =
\frac{1}{2 \sigma^{2}_{\text{cls}}} \, \mathcal{L}_{\text{focal}}
+
\frac{1}{2 \sigma^{2}_{\text{aux}}} \, \mathcal{L}_{\text{SCL}}
+
\frac{1}{2} \left( \log \sigma^{2}_{\text{cls}} + \log \sigma^{2}_{\text{aux}} \right),
\label{eq:ATCS_total_loss}
\end{equation}

where the final regularization term prevents the learned uncertainty parameters from diverging during optimization. This formulation allows the model to automatically balance the two loss components based on their relative uncertainty, rather than relying on a manually specified weighting factor.

In contrast, the MTS model was optimized using only the focal loss,
as monotonic survival behavior is inherently enforced by the cumulative structure of the MTS outputs (see equation \ref{eq:MTS_out}). Consequently, an explicit survival consistency loss was not required for this formulation.

\subsection{Cohort saliency analysis}

Image registration was performed solely for interpretability analysis and was not used during model training. 
The whole-body PET/CT volume of each patient was spatially
normalized to a common template space stratified by sex using the framework described by Jönsson et al. \cite{jonsson2022image}. 
The resulting deformation fields were used to transform individual saliency maps, generated by Grad-CAM 
\cite{selvaraju2017grad}
, into the template space.
Registered saliency maps were aggregated by voxel-wise averaging to obtain cohort-level representations, following the procedure described by Tarai et al. \cite{tarai2024prediction}.

Cohort-level saliency maps were computed separately for male and female patients for both the ATCS and MTS models at two predefined time intervals (6 months and 5 years).

\subsection{Experimentation}

A series of experiments was conducted to evaluate the proposed methods and to assess the effect of different modeling choices on survival prediction.
A 5-fold cross-validation strategy was employed, repeated with three different random seeds to ensure robustness. 
Performance was assessed at 6-months intervals over a 5-year horizon to maintain consistent temporal assessment. For each fold, time-specific AUC values were computed; these were then averaged across folds and seeds to obtain the time-specific overall AUC. Finally, the mean across all time points was used to derive the overall AUC. This evaluation protocol ensured stable estimates and enabled a fair comparison across methods.

\subsubsection{Comparison of Baseline and Proposed Methods}
As a first step, we compared the performance of the different methods (baseline/TCS, ATCS, MTS), all trained on the 13-channel PET/CT-Tumor input. This experiment provided a direct comparison with the baseline and allowed us to assess whether ATCS or the alternative modeling strategy MTS led to improved survival prediction performance.

\subsubsection{Effect of Imaging Input Modalities on MTS}

To assess the contribution of different imaging inputs, we conducted a series of ablation experiments using the MTS model with varying input configurations. These configurations included tumor-specific inputs derived from tumor segmentation, tissue-wise and original PET/CT projections, and combinations of tumor-focused and global imaging information. In addition, representative tumor slice inputs were evaluated in selected configurations. 
These configurations also varied in the number of input channels, reflecting different combinations of tissue-wise, tumor-masked, segmentation-based, and slice-based representations, as summarized in Table~\ref{tab:tissue-wise_exp_compact}.
These experiments were designed to examine whether integrating local tumor information with broader anatomical and metabolic context improves survival prediction performance within the MTS framework. 

\subsubsection{Effect of Temporal Resolution on MTS}
Finally, we examine the effect of varying the temporal resolution used for survival modeling. Models were trained with three different resolutions: 6-months, 1-month, and 10-days intervals. This analysis was performed to investigate whether finer time discretization provides improved short-term prediction accuracy and whether coarser resolutions lead to more stable long-term estimates.

\section{Results and Discussion}

\subsection{Results}

\subsubsection{Comparison of Baseline and Proposed Methods}
Table \ref{tab:main_method_comp} summarizes the comparison between the baseline/TCS, ATCS, and MTS methods, all trained on the 13-channel PET/CT-Tumor input (10 tissue-wise PET/CT projections, 2 tumor masked PET/CT projections, 1 tumor segmentation projection). Both ATCS and MTS achieved higher performance than the baseline. Overall AUC values were comparable between ATCS and MTS, although their time-specific behavior differed: ATCS performed better at earlier time points (up to 3 years), whereas MTS achieved superior performance at later time points (after 3 years). This complementary behavior suggests that ensembling or hybrid approaches may be beneficial in balancing short-term and long-term survival prediction.  

\begin{table}[!htbp]
\centering
\caption{Time-specific AUCs for overall survival (OS) prediction on the test set (n=292, events=84) using the DenseNet-121 backbone with 13-channel input (tissue-wise PET/CT projections, tumor masked PET/CT projections, tumor segmentation projections) using different methods (TCS, ATCS, and MTS). AUC values are reported at pre-specified time intervals from 6 months to 5 years, along with the mean AUC across all time intervals. Detailed mean $\pm$ standard deviation values are provided in the Supplementary Table \ref{tab:suppl_main_method_comp}.
The best performances are shown in bold.}
\label{tab:main_method_comp}
\begin{tabular}{p{3.0cm}cccccccccc@{\hspace{4mm}}c}
\hline
\multirow{2}{*}{Experiments} & \multicolumn{11}{c}{AUC (in years)}\\
\cline{2-12}
& .5 & 1 & 1.5 & 2 & 2.5 & 3  & 3.5  & 4 & 4.5 & 5 & Mean \\
\hline
Baseline/TCS & .742 & .735 & .778 & .771 & .778 & .765 & .755 & .780 & .772 & .790 & .767 \\
ATCS & \textbf{.751} & \textbf{.756} & \textbf{.814} & .791 & \textbf{.801} & \textbf{.809} & .792 & .801 & .805 & .822 & \textbf{.794} \\
MTS & \textbf{.751} & .722 & .791 & \textbf{.793} & .797 & .803 & \textbf{.794} & \textbf{.812} & \textbf{.821} & \textbf{.843} & .793 \\
\hline
\end{tabular}
\end{table}

\subsubsection{Effect of Imaging Input Modalities on MTS}

Table~\ref{tab:tissue-wise_exp_compact} summarizes the performance of the MTS model under different input configurations. Models combining tissue-wise PET/CT projections with tumor-specific information consistently achieved higher overall performance compared to configurations using either tumor-derived inputs or PET/CT projections alone.
Notably, the combination of tissue-wise PET/CT projections with tumor slice inputs ($Tissues + T_{slice}$) achieved strong performance across follow-up horizons, indicating that tumor slice representations can capture relevant prognostic information. 
However, incorporating additional tumor-derived inputs, including tumor-masked projections and segmentation-based projections, resulted in the highest overall performance, particularly in terms of mean AUC and long-term prediction. These findings suggest that while tumor slice representations provide a competitive alternative, combining multiple tumor representations with tissue-wise PET/CT information yields the most robust performance within the MTS framework.

\begin{table}[!htbp]
\centering
\caption{Ablation of input components for OS prediction using the MTS model trained with 13-channel input (tissue-wise PET/CT projections, tumor masked PET/CT projections, tumor segmentation projections). We report time-dependent AUCs at selected time intervals (0.5, 2.5, and 5 years) and the mean time-dependent AUC averaged across all pre-specified time intervals from 6 months to 5 years at 6 month intervals. 
The best performances are shown in bold.
$T_{slice}$: all-tumor slices from PET/CT, tumor slices overlayed from smallest to largest(on top); 
$Tissues$: PET/CT tissue-specific projections (original, bone, lean, adipose, air); 
$T_{masked}$: tumor-masked PET/CT projections; 
$T_{seg}$: tumor segmentation sum intensity projections(SIPs);}
\label{tab:tissue-wise_exp_compact}
\setlength{\tabcolsep}{.5pt}
\begin{tabular}{p{5.5cm}p{1.5cm}ccc@{\hspace{4mm}}c}
\hline
\multirow{2}{*}{Experiments} & \multirow{2}{*}{Channels} & \multicolumn{4}{c}{AUC (in years)}\\
\cline{3-6}
& & 0.5 & 2 & 5 & Mean \\
\hline

$T_{slice}$ & 2 & .696 & \textbf{.799} & .795 & .766 \\ 

$T_{masked}$ & 2 & .681 & .794 & .797 & .762 \\

$T_{masked}$ + $T_{seg}$ & 3 & .716 & .795 & .789 & .766 \\

$Tissues$ & 10 & .681 & .758 & .813 & .756 \\

$Tissues$ + $T_{slice}$ & 12 & .739 & .796 & .841 & .792 \\

$Tissues$ + $T_{masked}$ + $T_{seg}$ & 13 & .751 & .793 & \textbf{.843} & \textbf{.793} \\

$Tissues$ + $T_{masked}$ + $T_{seg}$ + $T_{slice}$ & 15 & \textbf{.752} & .798 & .838 & .792 \\

\hline
\end{tabular}

\end{table}

\subsubsection{Effect of Temporal Resolution on MTS}
Table \ref{tab:diff_time_res} presents results from models trained with different temporal resolutions. The 10-days resolution model achieved the highest mean AUC (0.799), showing strong performance in the first 1.5 years. However, the 6-months resolution model performed better during the later follow-up (2-5 years). 

This suggests a trade-off between fine-grained temporal modeling, suitable for short-term outcomes, and coarse-grained modeling for stable long-term prediction.

\begin{table}[!htbp]
\centering
\caption{Comparison of overall survival (OS) prediction performance of the MTS model trained with 13-channel input (tissue-wise PET/CT projections, tumor masked PET/CT projections, tumor segmentation projections) using different temporal resolutions (6-months, 1-month, and 10-days intervals). AUC values are reported at pre-specified time intervals from 6 months to 5 years, along
with the mean AUC across all time intervals. 
The best performances are shown in bold.}

\label{tab:diff_time_res}
\begin{tabular}{p{2.5cm}cccccccccc@{\hspace{4mm}}c}
\hline
\multirow{2}{*}{Experiments} & \multicolumn{11}{c}{AUC (in years)}\\
\cline{2-12}
& .5 & 1 & 1.5 & 2 & 2.5 & 3  & 3.5  & 4 & 4.5 & 5 & Mean \\
\hline
6 months & .751 & .722 & .791 & \textbf{.793} & \textbf{.797} & \textbf{.803} & \textbf{.794} & \textbf{.812} & \textbf{.821} & \textbf{.843} & .793 \\
1 month & .758 & .738 & .798 & .785 & .793 & .799 & .789 & .805 & .812 & .832 & .791 \\
10 days & \textbf{.774} & \textbf{.766} & \textbf{.817} & .788 & .794 & .802 & .789 & .806 & .818 & .832 & \textbf{.799} \\
\hline
\end{tabular}
\end{table}

\subsection{Discussion}
This study investigated two complementary temporal modeling strategies for PET/CT-based survival prediction in NSCLC: ATCS and MTS. 
While both approaches improved upon the baseline, they demonstrated distinct strengths across follow-up horizons, 
with ATCS performing better at earlier time points and MTS showing stronger performance at later time points. These findings highlight that temporal representation plays an important role in imaging-based prognostic modeling.

Our observations are consistent with prior PET/CT-based survival studies emphasizing the importance of multimodality fusion and combining local tumor information with global imaging context \cite{amini2021multi,ferrandez2023artificial,meng2023adamss}. Models integrating tissue-wise PET/CT projections with tumor-specific channels achieved higher predictive performance, indicating that tumor
representations and broader anatomical and metabolic context provide complementary prognostic information.

In contrast to earlier imaging-based survival studies that focused on single-time prediction \cite{chen2021deep} or Cox-based modeling \cite{katzman2018deepsurv}, the present work systematically compares two temporally structured deep learning formulations under identical PET/CT projection inputs, enabling direct assessment of how temporal parameterization influences survival prediction. While Cox-based and proportional hazards approaches assume a predefined functional relationship between covariates and hazard, more recent neural survival models, such as DeepHit \cite{lee2018deephit} and other discrete-time formulations, directly model event-time distributions. Our MTS formulation instead models conditional survival probabilities whose cumulative product explicitly enforces monotonic survival behavior. In parallel, the ATCS framework incorporates follow-up time as a conditioning variable within intermediate feature representations, enabling time-conditioned adaptation of imaging features rather than restricting temporal modeling to the output layer.

The observed temporal-resolution trade-off provides additional insight into cumulative survival modeling. In discrete survival formulations, finer time discretization increases the number of conditional predictions required to estimate long-term survival. While this allows more granular modeling of short-term survival dynamics, potentially improving early risk discrimination, it also increases the depth of probability propagation, making long-horizon estimates more sensitive to accumulated estimation variability.
In contrast, coarser discretization reduces the number of sequential conditional predictions, resulting in a shorter propagation chain from early to later time points. This can facilitate more stable transmission of early survival information to later intervals, leading to improved long-term performance. These findings indicate that temporal resolution influences both short-term sensitivity and long-term stability within cumulative survival modeling, and suggest potential value in exploring multi-resolution or adaptive time discretization strategies.

\begin{figure}[htb] 
  \centering 
  \includegraphics[width=.85\textwidth]{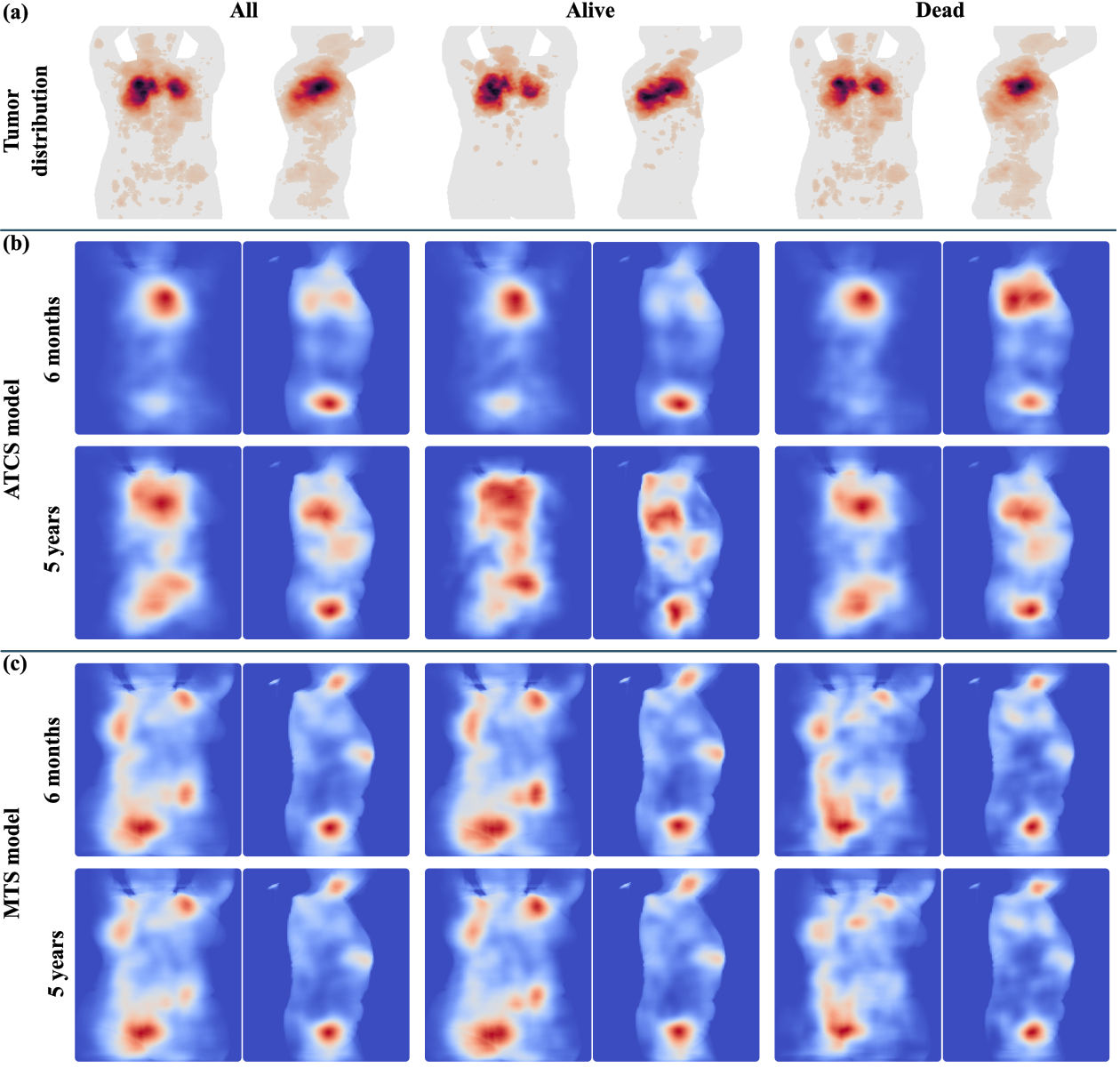} 
  \caption{Cohort saliency analysis for the female patient group. Cohort-averaged coronal and sagittal saliency maps are shown for (b) the ATCS model and (c) the MTS model at two representative follow-up time points (6 months and 5 years). Saliency maps are displayed for the full cohort (“All”), patients alive at follow-up (“Alive”), and patients deceased by follow-up (“Dead”). The corresponding tumor distribution maps are shown in (a) for reference. Saliency maps were generated using Grad-CAM and averaged across patients within each subgroup.}
  \label{fig:cohort_saliency_female}
\end{figure}

Cohort-level saliency analysis (see Fig. \ref{fig:cohort_saliency_female} for females and Supplementary Fig. \ref{fig:suppl_cohort_saliency_male} for males) further revealed differences in how temporal information is encoded. The MTS model exhibited relatively stable spatial importance patterns across horizons, reflecting its output-level temporal structure. 
In contrast, ATCS demonstrated more pronounced temporal variation in salient regions. Early predictions (e.g., 6 months) focused more on tumor-specific regions, whereas later predictions (e.g., 5 years) incorporated broader anatomical context, consistent with its conditioning of intermediate image representations on follow-up time.
This suggests that long-term survival prediction may rely not only on tumor-specific regions but also on broader anatomical or physiological context captured in baseline PET/CT imaging.

From a clinical perspective, generating survival estimates across multiple horizons from the baseline PET/CT image may support both early treatment decision-making and long-term follow-up planning. More broadly, the comparison of temporally structured modeling strategies illustrates that architectural design choices influence 
predictive performance.

Several limitations should be acknowledged. The study was conducted on a single dataset without external validation, which may limit generalizability. Additionally, the use of 2D projection representations, while computationally efficient, does not fully exploit volumetric spatial information. Future work may explore multi-center validation, integration of clinical variables (e.g., blood biomarkers and treatment), and segmentation-based tissue quantification. Extending the framework to 3D data and ensembling temporal strategies may further improve performance and clinical applicability.

\section{Conclusion}

In this study, we demonstrate that different temporal deep learning modeling strategies influence survival prediction from 2D PET/CT projections in patients with NSCLC. 
The complementary strengths of ATCS and MTS, for short‑term and long‑term horizons, respectively,  highlight the importance of temporal representation in imaging-based prognostic modeling.
Integrating tumor-specific and contextual imaging features further enhances predictive performance, supporting the value of combining local and global information. These findings suggest that careful architectural and temporal design choices can improve both discrimination and interpretability in survival modeling, providing a foundation for more flexible and clinically informative prognostic frameworks.

\begin{credits}
\subsubsection{\ackname} We would like to thank Hanna Jönsson who supported with the whole-body PET/CT image registration needed for the cohort saliency analysis. This study was supported by the Swedish Cancer Society (23 3123 Pj 01 H), Analytic Imaging Diagnostics Arena (AIDA), Lions Cancer Fund, and Makarna Eriksson Foundation. 

\subsubsection{\discintname}
Joel Kullberg and Håkan Ahlström are co-founders and employees of Antaros Medical. Antaros Medical holds the rights to patents related to image analysis methodology used for image registration. 
\end{credits}

\bibliographystyle{splncs04}
\bibliography{main_bibliography}

\newpage

\setcounter{section}{0}
\setcounter{table}{0}
\setcounter{figure}{0}

\renewcommand{\thetable}{S\arabic{table}}
\renewcommand{\thefigure}{S\arabic{figure}}
\renewcommand{\thesection}{S\arabic{section}}

\section{Additional Ablation Results}\label{section:suppl_additional_res}

We evaluated the impact of backbone architectures across different model variants and input configurations. The TCS model follows the configuration used in prior work \cite{tarai2026time}, where a 12-channel input was employed with a ResNet-50 backbone. To assess whether alternative backbones improve performance under this setting, we first conducted a backbone comparison for TCS (see Table \ref{tab:suppl_P1_arch_ablation}).
Based on these findings, DenseNet-121 was selected for subsequent experiments.

For the ATCS model, backbone performance was further evaluated using a 13-channel input configuration, including comparison with a vision transformer (ViT)-based architecture, where DenseNet-121 consistently achieved superior performance (see Table \ref{tab:suppl_P1-EN_arch_ablation}).


\begin{table}[!htbp]
\centering
\caption{Comparison of overall survival (OS) prediction performance with 12-channel input (tissue-wise PET/CT projections and tumor masked PET/CT projections) using different backbone architectures (ResNet-50, DenseNet-121, and Inception-v4) for the TCS model. AUC values are reported at pre-specified time intervals from 6 months to 5 years, along with the mean AUC across all horizons. 
The best performances are shown in bold.}

\label{tab:suppl_P1_arch_ablation}
\begin{tabular}{p{3.0cm}ccccccccccc}
\hline
\multirow{2}{*}{Experiments} & \multicolumn{11}{c}{AUC (in years)}\\
\cline{2-12}
& .5 & 1 & 1.5 & 2 & 2.5 & 3  & 3.5  & 4 & 4.5 & 5 & Mean \\
\hline
ResNet-50 & .694 & .710 & .761 & .760 & .754 & .747 & .742 & .762 & .756 & .772 & .746 \\
DenseNet-121 & .736 & .735 & \textbf{.779} & \textbf{.772} & \textbf{.779} & \textbf{.765} & \textbf{.760} & \textbf{.783} & \textbf{.772} & \textbf{.790} & \textbf{.767} \\
Inception-v4 & \textbf{.746} & \textbf{.736} & .765 & .756 & .757 & .751 & .744 & .764 & .754 & .766 & .754 \\
\hline
\end{tabular}
\end{table}


\begin{table}[!htbp]
\centering
\caption{Comparison of overall survival (OS) prediction performance with 13-channel input (tissue-wise PET/CT projections, tumor masked PET/CT projections, tumor segmentation projections) using different backbone architectures (DenseNet-121 and ViT) for the ATCS model. 
AUC values are reported at pre-specified time intervals from 6 months to 5 years, along with the mean AUC across all horizons. 
The best performances are shown in bold.}

\label{tab:suppl_P1-EN_arch_ablation}
\begin{tabular}{p{3.0cm}ccccccccccc}
\hline
\multirow{2}{*}{Experiments} & \multicolumn{11}{c}{AUC (in years)}\\
\cline{2-12}
& .5 & 1 & 1.5 & 2 & 2.5 & 3  & 3.5  & 4 & 4.5 & 5 & Mean \\
\hline
DenseNet-121 & \textbf{.751} & \textbf{.756} & \textbf{.814} & \textbf{.791} & \textbf{.801} & \textbf{.809} & \textbf{.792} & \textbf{.801} & \textbf{.805} & .822 & \textbf{.794} \\
ViT & .701 & .696 & .759 & .775 & .773 & .765 & .774 & .793 & .799 & \textbf{.828} & .766 \\
\hline
\end{tabular}
\end{table}

\newpage

\section{Extended results: main methods comparison}\label{section:core_arch_comparison}

\begin{table}[!htbp]
\centering
\caption{Time-specific AUCs (mean $\pm$ standard deviation) for overall survival (OS) prediction using the DenseNet-121 backbone with 13-channel input (tissue-wise PET/CT projections, tumor masked PET/CT projections, tumor segmentation projections) using different methods (TCS, ATCS, and MTS). AUC values are reported at pre-specified time intervals from 6 months to 5 years, along with the mean AUC across all time intervals.
The best performances are shown in bold.}
\label{tab:suppl_main_method_comp}
\resizebox{\textwidth}{!}{\begin{tabular}{p{2cm}cccccccccc@{\hspace{2mm}}c}
\hline
\multirow{2}{*}{Experiments} & \multicolumn{11}{c}{AUC (in years)}\\
\cline{2-12}
& .5 & 1 & 1.5 & 2 & 2.5 & 3  & 3.5  & 4 & 4.5 & 5 & Mean \\
\hline
Baseline/TCS & .742 {\scriptsize $\pm$ .030} & .735 {\scriptsize $\pm$ .028} & .778 {\scriptsize $\pm$ .035} & .771 {\scriptsize $\pm$ .037} & .778 {\scriptsize $\pm$ .037} & .765 {\scriptsize $\pm$ .039} & .755 {\scriptsize $\pm$ .033} & .780 {\scriptsize $\pm$ .033} & .772 {\scriptsize $\pm$ .039} & .790 {\scriptsize $\pm$ .035} & .767 {\scriptsize $\pm$ .016} \\
ATCS & \textbf{.751} {\scriptsize $\pm$ .029} & \textbf{.756} {\scriptsize $\pm$ .026} & \textbf{.814} {\scriptsize $\pm$ .029} & .791 {\scriptsize $\pm$ .019} & \textbf{.801} {\scriptsize $\pm$ .027} & \textbf{.809} {\scriptsize $\pm$ .027} & .792 {\scriptsize $\pm$ .031} & .801 {\scriptsize $\pm$ .029} & .805 {\scriptsize $\pm$ .036} & .822 {\scriptsize $\pm$ .039} & \textbf{.794} {\scriptsize $\pm$ .022} \\
MTS & \textbf{.751} {\scriptsize $\pm$ .031} & .722 {\scriptsize $\pm$ .025} & .791 {\scriptsize $\pm$ .026} & \textbf{.793} {\scriptsize $\pm$ .024} & .797 {\scriptsize $\pm$ .025} & .803 {\scriptsize $\pm$ .019} & \textbf{.794} {\scriptsize $\pm$ .018} & \textbf{.812} {\scriptsize $\pm$ .016} & \textbf{.821} {\scriptsize $\pm$ .017} & \textbf{.843} {\scriptsize $\pm$ .016} & .793 {\scriptsize $\pm$ .033} \\
\hline
\end{tabular}}
\end{table}

\newpage

\section{Effect of Training Strategies on ATCS}\label{section:suppl_ATCS_training_strat}

We evaluated the impact of training configurations on the ATCS model, including adaptive loss weighting (Wloss) and learning rate scheduling strategies (LR1: cosine with warmup, LR2: ReduceOnPlateau). Gradient accumulation was used across all configurations to enable stable training under memory constraints.

Table~\ref{tab:suppl_ATCS_diff_training_strats} summarizes the results. The combination of adaptive loss weighting and cosine scheduling (LR1 + Wloss) achieved the best performance across most time intervals and the highest mean AUC. Therefore, this configuration was used in the subsequent training. Configurations without adaptive loss weighting or using LR2 showed consistently lower performance.

\begin{table}[!htb]
\centering
\caption{
Effect of training configurations on overall survival (OS) prediction using the ATCS model trained with 13-channel input (tissue-wise PET/CT projections, tumor masked PET/CT projections, tumor segmentation projections).  
Wloss denotes adaptive uncertainty-based loss weighting; LR1 indicates cosine scheduling with warmup; LR2 indicates ReduceOnPlateau scheduling. 
AUC values are reported at pre-specified time intervals from 6 months to 5 years, along with the mean AUC across all horizons. 
The best performances are shown in bold.
}

\label{tab:suppl_ATCS_diff_training_strats}
\begin{tabular}{p{2.5cm}ccccccccccc}
\hline
\multirow{2}{*}{Experiments} & \multicolumn{11}{c}{AUC (in years)}\\
\cline{2-12}
& .5 & 1 & 1.5 & 2 & 2.5 & 3  & 3.5  & 4 & 4.5 & 5 & Mean \\
\hline
LR1 & .713 & .698 & .703 & .661 & .662 & .662 & .654 & .657 & .630 & .613 & .665 \\
LR1 + Wloss & \textbf{.751} & \textbf{.756} & \textbf{.814} & \textbf{.791} & \textbf{.801} & \textbf{.809} & \textbf{.792} & \textbf{.801} & \textbf{.805} & \textbf{.822} & \textbf{.794} \\
LR2 + Wloss & .650 & .683 & .715 & .702 & .687 & .687 & .678 & .683 & .679 & .697 & .686  \\
\hline
\end{tabular}
\end{table}

\newpage

\section{Effect of Training Strategies on MTS}\label{section:suppl_MTS_training_strat}

Tables \ref{tab:suppl_MTS_diff_LRSch_&_GA} and \ref{tab:suppl_MTS_diff_BS_LR_&_GA} summarize the impact of training strategies on MTS performance. Gradient accumulation consistently improved performance across learning rate schedulers, with the best results achieved using batch size = 6, learning rate = 1e-4, and GA = 4. Therefore, this configuration was used in the subsequent training. Without gradient accumulation, performance was consistently lower across most time intervals, indicating its importance for stable training under limited batch sizes.


\begin{table}[!htb]
\centering
\caption{Comparison of overall survival (OS) prediction performance of the MTS model trained with 13-channel input (tissue-wise PET/CT projections, tumor masked PET/CT projections, tumor segmentation projections) using different learning rate schedulers and gradient accumulation (GA) settings. Two schedulers were evaluated: cosine scheduler with warmup (LR1) and ReduceOnPlateau (LR2). AUC values are reported at pre-specified time intervals from 6 months to 5 years, along with the mean AUC across all horizons. 
The best performances are shown in bold.}

\label{tab:suppl_MTS_diff_LRSch_&_GA}
\begin{tabular}{p{2.5cm}ccccccccccc}
\hline
\multirow{2}{*}{Experiments} & \multicolumn{11}{c}{AUC (in years)}\\
\cline{2-12}
& .5 & 1 & 1.5 & 2 & 2.5 & 3  & 3.5  & 4 & 4.5 & 5 & Mean \\
\hline
LR1 & .707 & .692 & .768 & .782 & .784 & .785 & .781 & .797 & .799 & .824 & .772 \\
LR2 & .688 & .683 & .775 & .790 & .791 & .792 & .788 & .808 & .809 & .833 & .776 \\
LR1 + GA & \textbf{.751} & \textbf{.722} & \textbf{.791} & .793 & .797 & .803 & .794 & .812 & .821 & .843 & \textbf{.793} \\
LR2 + GA & .719 & .704 & \textbf{.791} & \textbf{.800} & \textbf{.805} & \textbf{.809} & \textbf{.801} & \textbf{.820} & \textbf{.828} & \textbf{.850} & \textbf{.793} \\

\hline
\end{tabular}
\end{table}


\begin{table}[!htb]
\centering
\caption{Effect of batch size, learning rate, and gradient accumulation (GA) on overall survival (OS) prediction performance using the MTS model trained with 13-channel input (tissue-wise PET/CT projections, tumor masked PET/CT projections, tumor segmentation projections). AUC values are reported at pre-specified time intervals from 6 months to 5 years, along with the mean AUC across all horizons. 
The best performances are shown in bold.}

\label{tab:suppl_MTS_diff_BS_LR_&_GA}
\begin{tabular}{p{1cm}p{1cm}p{.5cm}ccccccccccccc}
\hline
\multicolumn{3}{c}{Experiments} & \multirow{2}{*}{} & \multirow{2}{*}{} & \multicolumn{11}{c}{AUC (in years)}\\
\cline{1-3} \cline{6-16}
Batch & LR & GA & & & .5 & 1 & 1.5 & 2 & 2.5 & 3  & 3.5  & 4 & 4.5 & 5 & Mean \\
\hline
6 & 1e-4 & 4 & & & .751 & .722 & .791 & \textbf{.793} & \textbf{.797} & \textbf{.803} & \textbf{.794} & \textbf{.812} & \textbf{.821} & \textbf{.843} & \textbf{.793} \\
6 & 1e-4 & - & & & \textbf{.753} & \textbf{.728} & \textbf{.796} & .792 & .796 & .799 & .785 & .801 & .803 & .824 & .788 \\
24 & 4e-4 & - & & & .744 & .713 & .771 & .762 & .768 & .772 & .763 & .776 & .777 & .801 & .765 \\
\hline
\end{tabular}
\end{table}


\newpage 

\section{Cohort saliency analysis: Male}\label{section:suppl_cohort_saliency_male}

\begin{figure}[htbp] 
  \centering 
  \includegraphics[width=\textwidth]{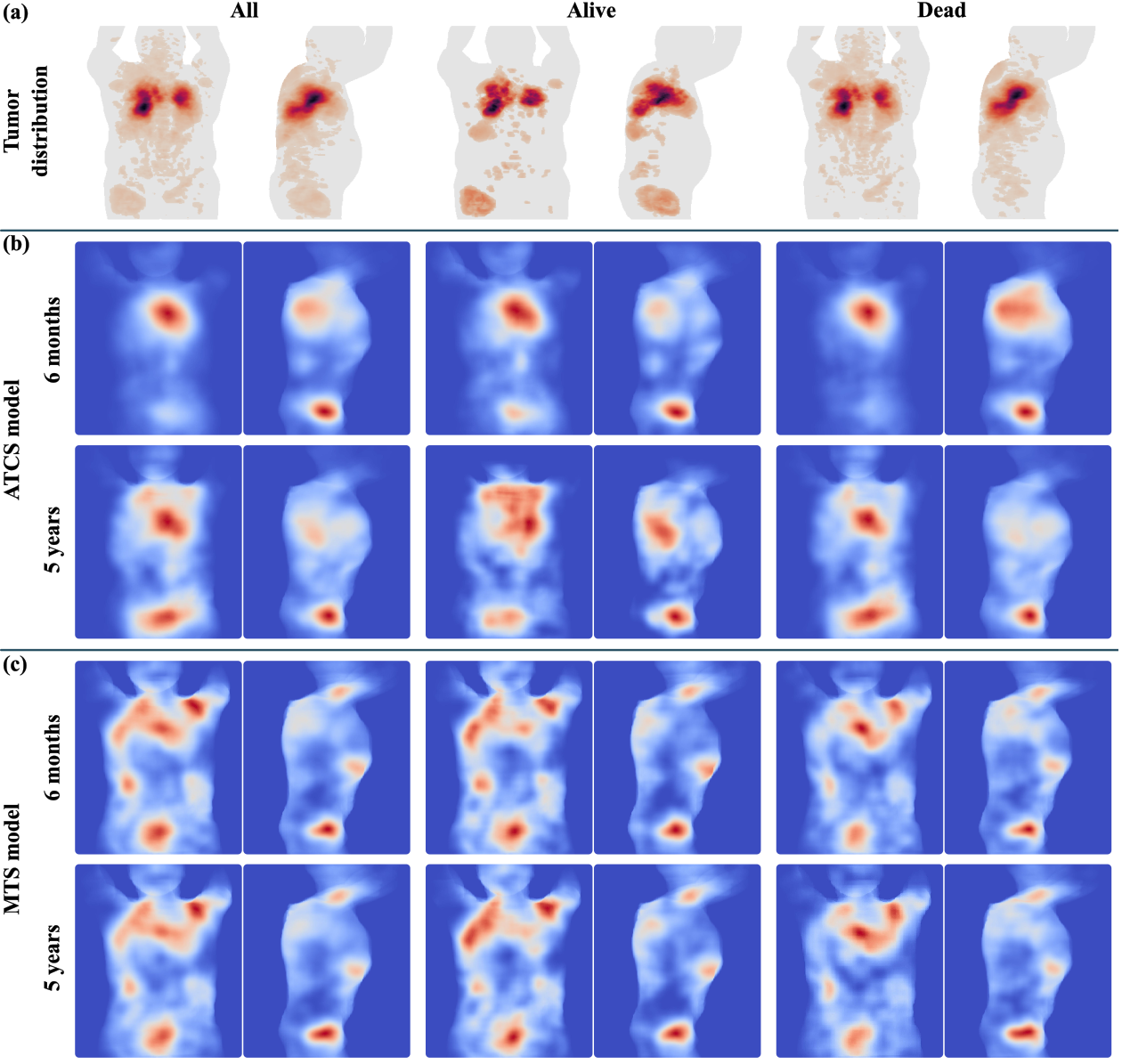} 
  \caption{Cohort saliency analysis for the male patient group. Cohort-averaged coronal and sagittal saliency maps are shown for (b) the ATCS model and (c) the MTS model at two representative follow-up time points (6 months and 5 years). Saliency maps are displayed for the full cohort (“All”), patients alive at follow-up (“Alive”), and patients deceased by follow-up (“Dead”). The corresponding tumor distribution maps are shown in (a) for reference. Saliency maps were generated using Grad-CAM and averaged across patients within each subgroup.}
  \label{fig:suppl_cohort_saliency_male}
\end{figure}

\end{document}